\documentclass[conference]{IEEEtran}
\IEEEoverridecommandlockouts
\usepackage{cite}
\usepackage{amsmath,amssymb,amsfonts}
\usepackage{algorithmic}
\usepackage{graphicx}
\usepackage{textcomp}
\usepackage{xcolor}
\usepackage{url}
\usepackage{amsmath}
\usepackage{amsfonts}
\usepackage{multirow}
\def\BibTeX{{\rm B\kern-.05em{\sc i\kern-.025em b}\kern-.08em
    T\kern-.1667em\lower.7ex\hbox{E}\kern-.125emX}}
\begin{document}

\title{Stochastic Graph Recurrent Neural Network\\}

\author{\IEEEauthorblockN{Tijin Yan}
\textit{Beijing Institute of Technology}\\
Beijing, China \\
\and
\IEEEauthorblockN{Hongwei Zhang}
\textit{Beijing Institute of Technology}\\
Beijing, China \\
\and
\IEEEauthorblockN{Zirui Li}
\textit{Beijing Institute of Technology}\\
Beijing, China \\
\and
\IEEEauthorblockN{Yuanqing Xia}
\textit{Beijing Institute of Technology}\\
Beijing, China \\
}

\maketitle

\begin{abstract}
Representation learning over graph structure data has been widely studied due to its wide application prospects. However, previous methods mainly focus on static graphs while many real-world graphs evolve over time. Modeling such evolution is important for predicting properties of unseen networks. To resolve this challenge, we propose SGRNN, a novel neural architecture that applies stochastic latent variables to simultaneously capture the evolution in node attributes and topology. Specifically, deterministic states are separated from stochastic states in the iterative process to suppress mutual interference. With semi-implicit variational inference integrated to SGRNN, a non-Gaussian variational distribution is proposed to help further improve the performance. In addition, to alleviate KL-vanishing problem in SGRNN, a simple and interpretable structure is proposed based on the lower bound of KL-divergence. Extensive experiments on real-world datasets demonstrate the effectiveness of the proposed model. Code is available at https://github.com/StochasticGRNN/SGRNN.
\end{abstract}
%

\section{Introduction}
Learning representations of nodes that encode non-Euclidean information to a low-dimensional space is challenging. Graph neural networks can well deal with non-Euclidean information and have been attracting much attention due to its ubiquitous applicability in variety domains such as social networks\cite{gu2017co}, intelligent transportation\cite{li2020social} and neural language processing\cite{yao2019graph}. Existing methods primarily focus on static graphs\cite{grover2016node2vec,armandpour2019robust}. However, many real-world networks are time dependent where nodes and edges may appear or disappear over time as shown in Fig 1. Green nodes and red nodes represent new and disappeared nodes and the solid and dashed lines represent original and new edges. It's important to model such evolution in representation learning over dynamic graphs.

Representation learning over dynamic graphs is a vast and interdisciplinary field. Researchers from different disciplines usually use modeling methods from their field. Here we divide them into three categories. 1) Tensor decomposition-based methods. Kazemi et al.\cite{kazemi2019relational} reviews the methods for tensor decomposition. Actually, it's analogous to matrix factorization methods \cite{zhu2016scalable,zhang2017timers} where time as an extra dimension is introduced. 2) Random walk-based methods. Random walk based models for dynamic graphs are usually extensions as random walk based embedding methods for static graph, i.e., \cite{yu2018netwalk,nguyen2018continuous}. Besides, Lambiotte\cite{lambiotte2016guide} review the methods of temporal random walks. 3) Deep learning-based methods. These methods can be divided into two categories. The first type integrates deep learning based time series models\cite{fraccaro2016sequential} with information aggregated from neighborhoods. And the other is based on temporal restricted Boltzmann machines, which are generative and probabilistic models and applied to link prediction tasks\cite{li2018restricted,divakaran2019temporal,li2014deep}. 

\begin{figure}[t]
	\centering
	\includegraphics[width=0.46\textwidth, height=0.13\textheight]{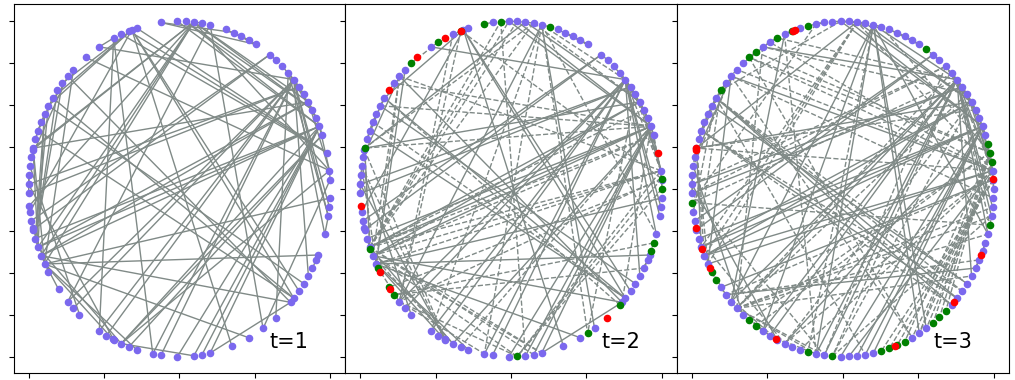} 
	\caption{Dynamic graph}
	\label{Dyn_graph}
\end{figure}

Although many works have been proposed for representation learning over dynamic graphs, they have some limitations. Firstly, some methods\cite{zhou2018dynamic,goyal2018dyngem} can not model long-range temporal dependencies. Besides, many works cannot capture changes in node attributes and graph topology at the same time. In addition, some methods\cite{seo2018structured,trivedi2018dyrep} cannot model the deletion of node or edges. Last but not least, many methods cannot model the uncertainty in latent representation, which is proved efficient for node embedding methods in VGRNN. However, VGRNN incorporates stochastic states to update recurrent states, which limits its expressive power. 

To tackling the problem mentioned above, we propose stochastic graph recurrent neural network (SGRNN) for the representation learning over dynamic graphs. Specifically, GRNN and VAE are integrated together to simultaneously model temporal dependencies and uncertainty in latent representations. Besides, deterministic states are separated from stochastic states in the iterative process in order to suppress mutual interference. Additionally, the assumption that latent representations follow Gaussian distribution is inappropriate for complex networks. Therefore, in order to further improve the performance of SGRNN, semi-implicit variational inference (SIVI) that introduce flexible variational distributions is integrated into the framework of SGRNN. In the end, in order to alleviate posterior collapse problem in SGRNN, a simple but interpretable structure for inference network is proposed based on the lower bound of KL-divergence. And experiments on six real-world datasets demonstrate that SGRNN and SI-SGRNN outperforms existing methods over dynamic link prediction.

\section{Background and preliminaries}
\subsubsection{VAE \& VGAE} VAE\cite{kingma2013auto} is a deep Bayesian network which models the relationship between two random variables x and z. Then $p_{\theta}(x) = \mathbb{E}_{p(z)}p_{\theta}(x|z)$ can be used for density estimation. VAE introduces an approximation posterior $q(z|x)$ to estimate true posterior distribution $p(z|x)$ that is intractable. And the marginal log-likelihood can be formulated as 
\begin{equation}
\log p(x) = \mathcal{L}(\theta, \phi; x) + D_{KL}[q(z|x)|p(z|x)]
\end{equation}
where the evidence lower bound $\mathcal{L}$ is
\begin{equation}\label{lower_bound}
\begin{aligned}
\mathcal{L}(\theta, \phi; x) &= -\mathbb{E}_{q(z|x)}[-\log q(z|x) + \log p(x,z)]\\
\end{aligned}
\end{equation}

VGAE\cite{kipf2016variational} is the first work that combines VAE and GNN together for link prediction task based on static graph structure data. Besides, it proved that incorporate node attributes and graph topology together can significantly improve the performance of model.

\subsubsection{GRNN} The combination of GNN and autoregressive models has been explored by many works. GCRN\cite{seo2018structured} is proposed to model time series data over a static graph. However, it can not model sequences with high variations. VGRNN\cite{hajiramezanali2019variational} extends it to dynamic graph and proposes GRNN which model node deletion as removing all edges connected to it.  WD-GCN\cite{manessi2020dynamic} combines spectral GCN with LSTM. RgCNN\cite{narayan2018learning} integrates spatial GCN and LSTM together to model dynamic graph. DyGGNN\cite{taheri2019learning} combines gated graph sequence neural network(GGNN) and LSTM while DySAT\cite{sankar2020dysat} combines graph attention neural network(GAT) with self-attention module in Transformer\cite{vaswani2017attention}. However, almost all the methods mentioned above are deterministic modeling without introducing stochastic latent variables to model uncertainty. 

VGRNN integrates VRNN and GCN together to model the uncertainty in latent representations and proves its effectiveness through experiments on some real-world datasets. 
However, deterministic states and stochastic states interfere with each other in the iterative process of VGRNN. Therefore, to further improve the model expression power, deterministic states are separated from stochastic states in SGRNN. 


\subsubsection{KL-vanishing} When paired with strong auto regressive model, VAE suffers from a well known problem named KL-\textit{vanishing} or \textit{posterior collapse} problem. Actually, many works have been proposed to prevent posterior collapse and can be divided into following categories: 1) Fix KL. Fixing KL as a positive constant is an intuitive method to prevent KL-vanishing\cite{davidson2018hyperspherical,guu2018generating,xu2018spherical,razavi2019preventing}. However, these approaches force the same constant KL and lose flexibility to allow various KLs for different data points. 2) Change distributions. The motivation for such methods is that Gaussian distribution is actually inappropriate for latent distributions. And many works \cite{van2017neural,davidson2018hyperspherical,jimenez2015variational} have proved that flexible latent distributions will result in better posterior approximation. Actually SIVI also proposes a more flexible distribution for inference. 3)View of optimization. Some recent works analyze the problem from a view of optimization. \cite{he2019lagging} adds additional train loops for the inference network to solve the lagging in inference. \cite{li2019surprisingly} proposes to initialize the inference network with an encoder pretrained from an AE objective. However, the computational costs of these methods are usually high.

In this paper, a simple and interpretable structure that can be implemented with simply one line code modification is proposed to prevent KL-vanishing problem. Specifically, the lower bound of KL-divergence is derived and can be controlled with a batch normalization. Besides, it uses the information of prior distribution to guide the generation of the mean value of posterior distribution. Finally, it can be easily extended to prevent KL-vanishing problem for sequential VAE models. More details can be found in Section \ref{sec_KL_a}.

\section{Methodology}\label{method}
In this section, detail components of SGRNN will be introduced. Besides, the method of integrating SIVI to the framework of SGRNN will be discussed. In addition, the lower bound of KL divergence and the simple but interpretable structure guided by the lower bound will be introduced.
\subsection{Notations}

For a dynamic graph $\mathcal{G} = \{G^{(1)}, G^{(2)}, \cdots, G^{(T)}\}$, $G^{(t)} =(\mathcal{V}^{(t)}, \mathcal{E}^{(t)})$ is the graph at time $t$, $\mathcal{V}^{(t)}$ and $\mathcal{E}^{(t)}$ are corresponding nodes and edges.
Assume $N_t$ is the number of nodes at time $t$, $M$ is the dimension of node attributes, which is constant across time. Besides, denote the sequence of adjacency matrix as $\mathcal{A} = \{\textbf{A}_{1}, \textbf{A}_{2}, \cdots, \textbf{A}_{T}\}$ and the node attributes sequence as $\mathcal{X} = \{\textbf{X}_{1}, \textbf{X}_{2}, \cdots, \textbf{X}_{T}\}$. Note that $\textbf{A}_{t}$ and $\textbf{X}_{t}$ are matrices with shape $N_t\times N_t$ and $N_t\times M$. 

\subsection{SGRNN}
SGRNN models node deletion as removing all edges connected to it. Besides, it can capture both of the topological evolution and node attributes changes. And the structure of SGRNN can be divided into following three components: prior network, generation network and inference network. 
\subsubsection{Prior Distribution} 
\begin{figure}[t]
	\centering
	\includegraphics[width=0.20\textwidth, height=0.17\textheight]{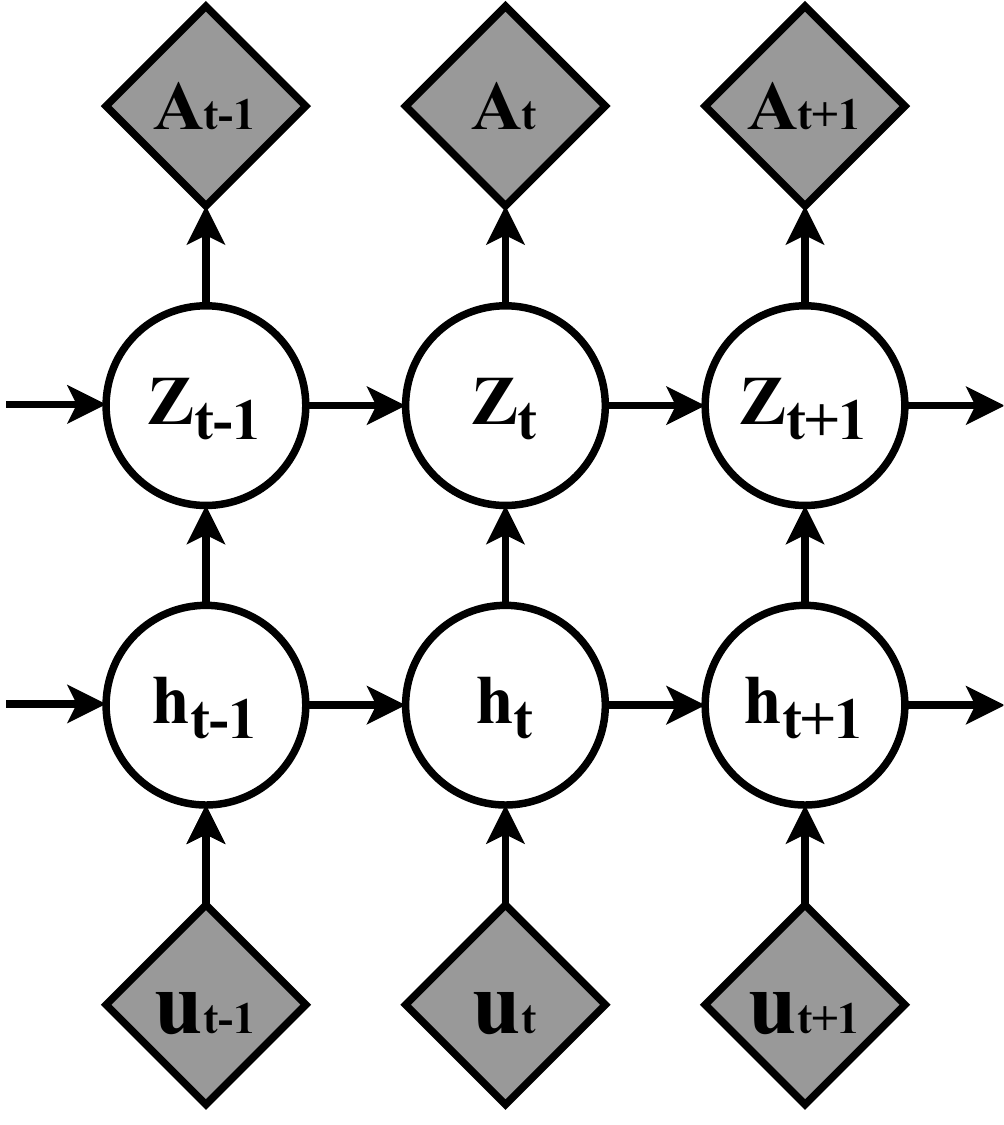} 
	\caption{Generative model $p_{\theta}$}
	\label{Gen_fig1}
\end{figure}

SGRNN is defined as a sequential generative model $p_{\theta}$ by temporally combining VGAE and GRNN together, as shown in Fig \ref{Gen_fig1}. Different from VGRNN, the latent variable $\textbf{Z}_t$ directly depends on $\textbf{Z}_{t-1}$ by $p_{\theta_z}(\textbf{Z}_t|\textbf{Z}_{t-1}, \textbf{h}_{t})$, as it does in a state space model (SSM). Compared with VGRNN, it separates deterministic and stochastic parts of the prior distribution and make hidden state $\textbf{h}_t$ entirely deterministic instead of depending on past latent variables $\textbf{Z}$. In addition, the split makes the structure of posterior distribution approximation easier to construct where $\textbf{Z}_t$ still satisfy Markov property. Assume the initial latent variable $\textbf{Z}_0=\textbf{0}$ and hidden state $\textbf{h}_0=\textbf{0}$. The hidden states $\textbf{h}_{1:T}$ are determined by $\textbf{h}_0$, $\textbf{u}_{1:T}$ though recursion $\textbf{h}_{t} = f_{\theta_d}(\textbf{h}_{t-1},  \textbf{u}_t)$, where $f_{\theta_d}$ is a nonlinear transformation. $f_{\theta_d}$ is implemented with a MLP in the experiments. Besides, $\textbf{u}_t$ represents the input information to the model, which depends on the type of task. 

In this paper, $\textbf{u}_t$ is set as $\{\textbf{A}_t, \textbf{X}_t\}$ for detecting unobserved links in $G^{(t)}$ at the current moment,  $\{\textbf{A}_{t-1}, \textbf{X}_{t-1}\}$ for link prediction in the next snapshot $G^{(t)}$. All in all, the prior distribution and hidden states can be factorized as
\begin{equation}
\begin{aligned}
\ \ p_{\theta}(\textbf{Z}_{1:T}&, \textbf{h}_{1:T}|\textbf{u}_{1:T}, \textbf{Z}_0, \textbf{h}_0)\\
=& \prod_{t=1}^Tp_{\theta_z}(\textbf{Z}_t|\textbf{Z}_{t-1}, \textbf{h}_t)p_{\theta_h}(\textbf{h}_t|\textbf{h}_{t-1}, \textbf{u}_t)
\end{aligned}
\end{equation}

The prior transition distribution $p_{\theta_z}(\textbf{Z}_t|\textbf{Z}_{t-1}, \textbf{h}_t)$ is assumed to be a Gaussian distribution with diagonal covariance matrix $\mathcal{N}(\textbf{Z}_t; \boldsymbol{\mu}_t, \boldsymbol{\sigma}_t)$, whose mean and standard deviation are determined by a parameterized neural network that depends on $\textbf{Z}_{t-1}$ and $\textbf{h}_{t}$.
\begin{equation}
\begin{aligned}
\boldsymbol{\mu}_t^{(p)} &= \text{NN}_1^{(p)}(\textbf{Z}_{t-1}, \textbf{h}_t)\\
\boldsymbol{\sigma}_t^{(p)} &= \text{SoftPlus}(\text{NN}_2^{(p)}(\textbf{Z}_{t-1}, \textbf{h}_t))\\
\end{aligned}
\end{equation}
\subsubsection{Generation} As shown in Fig \ref{Gen_fig1}, the generative distribution depends on $\textbf{Z}_t$. Assume $p_{\theta}(\textbf{A}_t | \textbf{Z}_t) \sim \text{Bernoulli}(\pi_t)$, where 
$\pi_t = \psi_{\text{dec}}(\textbf{Z}_t)$.
Actually $\psi_{\text{dec}}$ can be highly flexible functions of any form. In the implementation, an inner-product decoder is adopted, which is the same as that in VGAE. Specifically, assume $\textbf{A}_t^{(i,j)}$ are the elements of adjacency matrix $\textbf{A}_t$, $\sigma(\cdot)$ is the logistic sigmoid function. The transformation can be formulated as 
\begin{equation}
\begin{aligned}
p(\textbf{A}_t|\textbf{Z}_t) &= \prod_{i=1}^{N_t}\prod_{j=1}^{N_t}p(\textbf{A}^{(i,j)}_t|\textbf{Z}_t^{(i)}, \textbf{Z}^{(j)}_{t})\\
\end{aligned}
\end{equation}
where 
$p(\textbf{A}_{t}^{(i,j)} =1 |\textbf{Z}_t^{(i)}, \textbf{Z}_t^{(j)}) = \sigma({\textbf{Z}_t^{(i)}}^T\textbf{Z}_t^{(j)})$. 

\subsubsection{Inference} The same deterministic states are adopted in the inference network as those of generative model, that is $q(\textbf{h}_{1:T}|\textbf{u}_{1:T},\textbf{h}_0)=p_{\theta_h}(\textbf{h}_{1:T}|\textbf{u}_{1:T},\textbf{h}_0)$. And the posterior distribution can be factorized as 
\begin{equation}
\begin{aligned}
q_{\phi}(\textbf{Z}_{1:T}&, \textbf{h}_{1:T}|\textbf{u}_{1:T}, \textbf{A}_{1:T}, \textbf{Z}_0, \textbf{h}_0)\\
=& q_{\phi}(\textbf{Z}_{1:T}|\textbf{A}_{1:T}, \textbf{h}_{1:T}, \textbf{Z}_0) q(\textbf{h}_{1:T}|\textbf{u}_{1:T},\textbf{h}_0)
\end{aligned}
\end{equation}

From Fig \ref{Gen_fig1}, given the input information $\textbf{u}_{1:T}$, output $\textbf{A}_{1:T}$, initial deterministic state $\textbf{h}_{0}$ and initial stochastic latent variable $\textbf{Z}_0$, the true posterior distribution can be factorized as Eq. (\ref{d_sep}) according to \textit{d-separation}\cite{geiger1990identifying}.
\begin{small}
	\begin{equation}\label{d_sep}
	p_{\theta}(\textbf{Z}_{1:T}|\textbf{h}_{1:T}, \textbf{A}_{1:T}, \textbf{Z}_{0}, \textbf{h}_0) = \prod_{t=1}^T p_{\theta}(\textbf{Z}_{t}|\textbf{Z}_{t-1}, \textbf{h}_{t:T}, \textbf{A}_{t:T})
	\end{equation}
\end{small}
\begin{figure}[t]
	\centering
	\includegraphics[width=0.46	\textwidth, height=0.24\textheight]{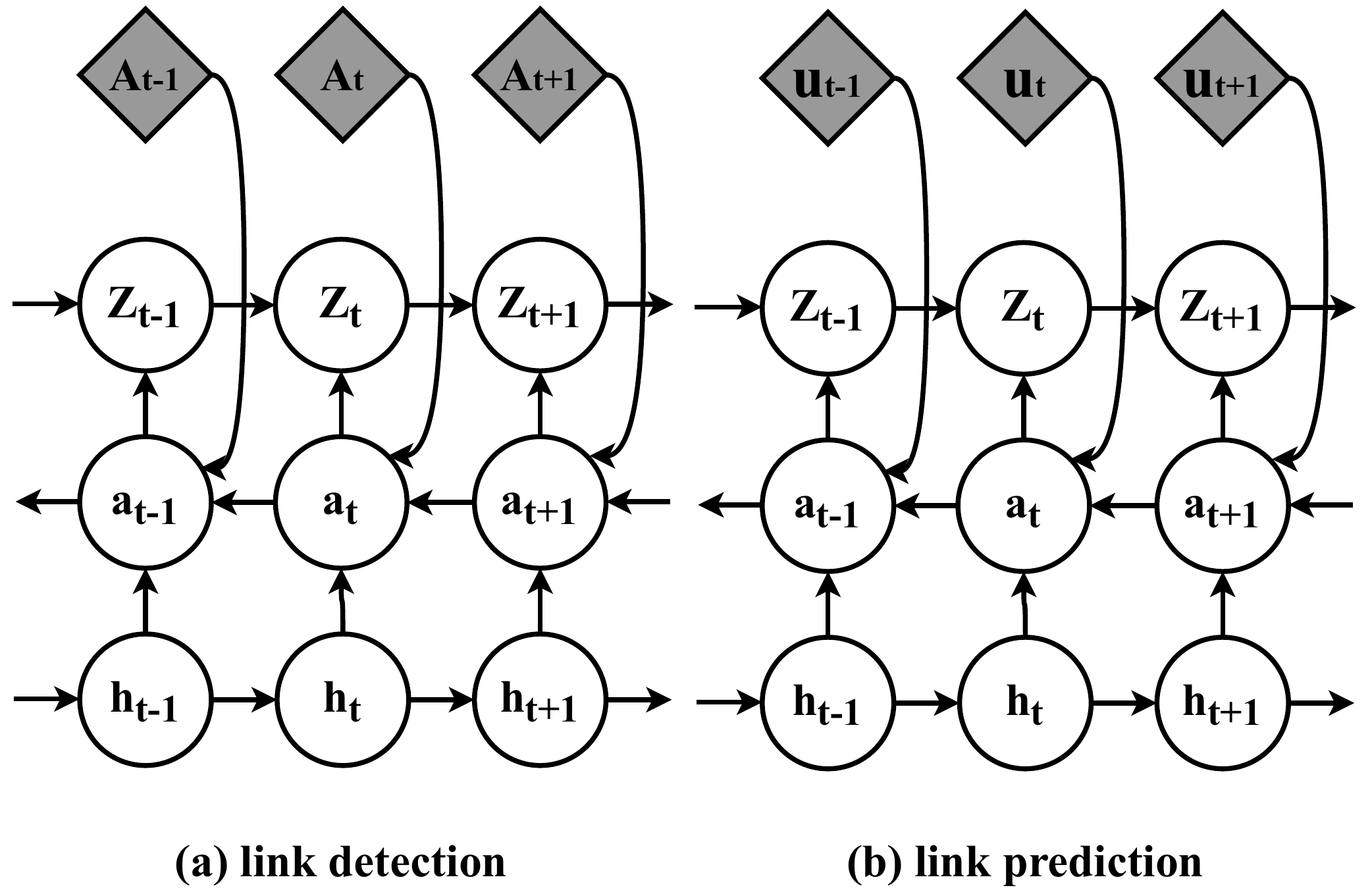} 
	\caption{Inference model $q_{\phi}$}
	\label{Inf_fig2}
\end{figure}

It can be known from Eq. (\ref{d_sep}) that $\textbf{Z}_t$ depends on the present and future reconstruction outputs $\textbf{A}$ and hidden states $\textbf{h}$ instead of the past ones when given $\textbf{Z}_{t-1}$. Therefore, this structure is also adopted for posterior distribution approximation as shown in Eq. (\ref{pos_ap}) instead of a mean-field approximation across time steps. In summary, the posterior distribution directly depends on $\textbf{A}_{t:T}$ and $\textbf{h}_{t:T}$ while depending on the past ones through $\textbf{Z}_{t-1}$. And the structure of inference network for unobserved link detection is shown in Fig \ref{Inf_fig2} (a).
\begin{small}
	\begin{equation}\label{pos_ap}
	\begin{aligned}
	&q_{\phi}(\textbf{Z}_{1:T}|\textbf{h}_{1:T}, \textbf{A}_{1:T}, \textbf{X}_{1:T}) =\prod_{t=1}^T q_{\phi}(\textbf{Z}_t|\textbf{Z}_{t-1}, \textbf{h}_{t:T}, \textbf{A}_{t:T})\\
	=&\prod_{t=1}^T q_{\phi_z}(\textbf{Z}_t|\textbf{Z}_{t-1}, \textbf{a}_t = g_{\phi_a}(\textbf{a}_{t+1}, \textbf{h}_t,\textbf{A}_t))\\
	\end{aligned}
	\end{equation}
\end{small}
where $g_{\phi_a}$ represents a GRNN cell at each time step. Additionally, the architecture is fine tuned to prevent information leakage for link prediction task, which is shown in Fig \ref{Inf_fig2} (b). Similarly to prior distribution, $q_{\phi_z}(\textbf{Z}_t|\textbf{Z}_{t-1}, \textbf{a}_t)$ is set as a Gaussian distribution with diagonal covariance, whose mean and standard deviation are determined by a parameterized neural network.
\begin{equation}\label{post_para}
\begin{aligned}
\boldsymbol{\mu}_t^{(q)} &= \text{NN}_1^{(q)}(\textbf{Z}_{t-1}, \textbf{a}_t)\\
\boldsymbol{\sigma}_t^{(q)} &= \text{SoftPlus}(\text{NN}_2^{(q)}(\textbf{Z}_{t-1}, \textbf{a}_t))\\
\end{aligned}
\end{equation}

\subsubsection{Loss} The graph at each time step corresponds to a VGAE. And all the intractable variables are estimated by sampling. Besides, the latent variables $\textbf{Z}$ cannot be integrated analytically.
Therefore, reparameterization trick is used to estimate the evidence lower bound (ELBO). In the end, the ELBO $\mathcal{L}$ for SGRNN is
\begin{equation}\label{loss}
\begin{aligned}
\mathcal{L} = \sum^T_{i=1}&\big\{ \mathbb{E}_{\textbf{Z}_{t}\sim q_{\phi}(\textbf{Z}_{t}|\textbf{a}_{t}, \textbf{Z}_{t-1})}[\log p_{\theta}(\textbf{A}_{t}|\textbf{Z}_{t})] \\
-& \text{KL}(q_{\phi}(\textbf{Z}_{t}|\textbf{a}_{t}, \textbf{Z}_{t-1}) || p_{\theta}(\textbf{Z}_{t}|\textbf{Z}_{t-1}, \textbf{h}_t))\big\}
\end{aligned}
\end{equation}

The parameters of generative network $\theta$ and inference network $\phi$ are joint optimized by the ELBO. SGRNN can deal with long time dependency through a nonlinear SSM structure and is compatible with graphs of different sizes and structures at each snapshot. Compared to GRNN, SGRNN can also model the uncertainty of the latent variables, which improves the robustness of the model.

\subsection{Improve the posterior approximation} \label{sec_KL_a}
In the experiments, the posterior parameterization method introduced in Eq. (\ref{post_para}) will result in a small value in KL term 
in Eq. (\ref{loss}). The ELBO will degenerate into reconstruction loss, which is the same as normal autoencoder model. KL-vanishing will severely affects the performance of the model. And the reason for KL-vanishing may be that $g_{\phi_a}$ learns focuses on hidden states $\textbf{h}$ instead of the information from present and future in $\textbf{a}$. 

Actually, it has been proved that when VAE is trained with strong auto regressive model, there will usually be a KL-vanishing problem. And many works have been proposed to solve the KL-vanishing problem so far. Inspired by \cite{zhu2020batch}, we propose a method based on batch normalization for posterior collapse problem in sequential VAE.

As the prior and posterior distribution are assumed to be a Gaussian distribution, the KL term between $q_{\phi}(\textbf{Z}|\textbf{X})$ and $\textbf{p}_{\theta}(\textbf{Z})$ at each time step can be formulated as:
\begin{small}
	\begin{equation}
	\begin{aligned}
	&\text{KL}(\mathcal{N}(\boldsymbol{\mu}_q, \boldsymbol{\sigma}_q)|\mathcal{N}(\boldsymbol{\mu}_p, \boldsymbol{\sigma}_p)) \\
	=& \frac{1}{2b}\sum^b_{j=1}\sum^d_{i=1}\left\{\frac{{\boldsymbol{\sigma}_q^{(i,j)}}^2}{{\boldsymbol{\sigma}_p^{(i,j)}}^2} + \frac{(\boldsymbol{\mu}_p^{(i,j)}-\boldsymbol{\mu}_q^{(i,j)})^2}{{\boldsymbol{\sigma}_p^{(i,j)}}^2} - \log \frac{{\boldsymbol{\sigma}_q^{(i,j)}}^2}{{\boldsymbol{\sigma}_p^{(i,j)}}^2}-1 \right\}\\
	\end{aligned}
	\end{equation}
\end{small}
where $d$ is the dimension of latent variables, $b$ is the number of edges. It is obvious that when $b$ gets larger, the KL term will approach its expectation. We can consider $\sum^b_{j=1}{\mu_q^{(i,j)}}^2/b$ as sample mean of ${\mu_q^{(i)}}^2$. And the other terms in $\{\boldsymbol{\cdot} \}$ can also be seen as sample mean terms. As $e^{x} \ge x+1$, the expectation of KL term can be simplified as
\begin{small}
	\begin{equation}\label{KL}
	\mathbb{E}(\text{KL}(\mathcal{N}(\boldsymbol{\mu}_q, \boldsymbol{\sigma}_q)|\mathcal{N}(\boldsymbol{\mu}_p, \boldsymbol{\sigma}_p))) \ge \frac{1}{2}\sum^d_{i=1}\mathbb{E}\left\{\frac{(\boldsymbol{\mu}_q^{(i)} -\boldsymbol{\mu}_p^{(i)})^2}{{\boldsymbol{\sigma}_p^{(i)}}^2}\right\}
	\end{equation}
\end{small}

Assume 
$l_i = (\boldsymbol{\mu}_q^{(i)} -\boldsymbol{\mu}_p^{(i)})/(\boldsymbol{\sigma}_p^{(i)})$. Then the expectation can be derived as $\mathbb{E}(l_i^2) = \mu(l_i)^2 + \sigma(l_i)^2$. A parameter-fixed batch normalization\cite{ioffe2015batch} is adopted to control the mean and standard deviation of $l_i$.
\begin{equation}
\hat{l}_i = \gamma * \frac{l_i - \boldsymbol{\mu}_{\mathcal{B}i}}{\boldsymbol{\sigma}_{\mathcal{B}i}} + \beta 
\end{equation}
where $\boldsymbol{\mu_{\mathcal{B}i}}$ is the mean of $l_i$, $\boldsymbol{\sigma_{\mathcal{B}i}}$ is the standard deviation of $l_i$. $\gamma$ and $\beta$ are the scale and shift parameters. In order to control the expectation of $l_i$, $\gamma$ is freezed to form a parameter-fixed BN. And $\beta$ is leaved as a learnable parameter that makes the lower bound more flexible. The lower bound of the KL divergence can be controlled by $\gamma$  With the parameter-fixed BN, we can derive the new lower bound of KL term as 
\begin{small}
	\begin{equation}
	\begin{aligned}
	\mathbb{E}(\text{KL}) &\ge \frac{1}{2}\sum^d_{i=1}\mathbb{E}\left\{\frac{(\boldsymbol{\mu}_q^{(i)} -\boldsymbol{\mu}_p^{(i)})^2}{{\boldsymbol{\sigma}_p^{(i)}}^2}\right\}
	=\frac{d(\gamma^2 + \beta^2)}{2}
	\end{aligned}
	\end{equation}
\end{small}
Then the new structure can be derived as
\begin{equation}\label{stru}
\boldsymbol{\mu}_q = \boldsymbol{\mu}_p + \boldsymbol{\sigma}_p * \text{FixedBN}(\text{NN}_1^{(q)}(\textbf{Z}_{t-1}, \textbf{a}_t))
\end{equation}

The structure indicates that if the information of prior distribution is used to guide the generation of mean value of posterior distribution, the lower bound of KL divergence can be controlled by $\gamma$. Besides, we also explore two variants. In the first variant, $\sigma$ is assumed to be a constant \textbf{1}, that is $\boldsymbol{\mu}_q = \boldsymbol{\mu}_p + \text{FixedBN}(\text{NN}_1^{(q)}(\textbf{Z}_{t-1}, \textbf{a}_t))$. The other is to directly optimize the difference between prior mean and posterior mean, that is $\boldsymbol{\mu}_q = \boldsymbol{\mu}_p + \text{NN}_1^{(q)}(\textbf{Z}_{t-1}, \textbf{a}_t))$. The detail comparative study can be found in section \ref{sec_kl}.

\subsection{SI-SGRNN}
The assumption that the latent representations follow Gaussian distribution is inappropriate. Therefore, SIVI is introduced and it expands the variational family to incorporate highly flexible variational distributions and can approach ELBO asymptotically via a lower bound. 

The prior distribution and generative network remain unchanged. Assume $ \textbf{Z}_{t}\sim q(\textbf{Z}_{t}|\boldsymbol{\psi}_t)$, $\boldsymbol{\psi}_t\sim q_{\phi}(\boldsymbol{\psi}_t|\textbf{Z}_{t-1}, \textbf{a}_{t})$. Note that $q(\textbf{Z}_{t}|\boldsymbol{\psi}_t)$ is implemented with a Gaussian distribution while $\boldsymbol{\psi}_t$ could be more flexible. A random noise $\boldsymbol{\epsilon}_{t}$ is introduced in $\boldsymbol{\psi}_t$ to help construct a more flexible inference model. The inference network is modified as 
\begin{small}
	\begin{equation}
	\begin{aligned}
	\textbf{r}_t^{(0)}&=\textbf{a}_t~~~	\textbf{r}_t^{(j)} = \text{GNN}_j(\textbf{A}_t, \text{CONCAT}(\textbf{Z}_{t-1}, \textbf{r}_t^{(j-1)}, \boldsymbol{\epsilon}_t))\\
	\boldsymbol{\mu}_{t} &= \text{GNN}_{\mu}(\textbf{A}_t, \textbf{r}_t^{(L)})~~~	\boldsymbol{\sigma}_{t} = \text{SoftPlus}(\text{GNN}_{\sigma}(\textbf{A}_t, \textbf{r}_t^{(L)}))\\
	\end{aligned}
	\end{equation}
\end{small}
where $j =1, \cdots,L$, $L$ is the number of stochastic layers, \text{GNN} represents a general graph neural network layer.

\subsubsection{Loss} Similar to SGRNN, the ELBO of this new structure can be derived as
\begin{equation}
\begin{aligned}
\mathcal{L}=-\sum^T_{t=1}& \textbf{KL}(\mathbb{E}_{\psi_t\sim q_{\phi}(\psi_t|\textbf{Z}_{t-1}, \textbf{a}_t)}q(\textbf{Z}_t|\psi_t)|p(\textbf{Z}_t|\textbf{Z}_{t-1},\textbf{h}_{t}))\\
+& \mathbb{E}_{\psi_t\sim q_{\phi}(\psi_t|\textbf{Z}_{t-1}, \textbf{a}_t)} \mathbb{E}_{\textbf{Z}_t\sim q(\textbf{Z}_t|\psi_t)} \log p(\textbf{A}_t|\textbf{Z}_t)\\
\end{aligned}
\end{equation}

Since direct direct optimization of $\mathcal{L}$ is not tractable in SIVI, a lower bound of ELBO is derived based on Jensen's inequality as Eq. (\ref{new_l}). And detail derivation process can be found in supplements.
\begin{equation}\label{new_l}
\begin{aligned}
\mathcal{\underline{L}} =& - \sum^T_{i=1}\mathbb{E}_{\psi_t\sim q_{\phi}(\psi_t|\textbf{Z}_{t-1}, \textbf{a}_t)} \textbf{KL}(q(\textbf{Z}_t|\psi_t)| p(\textbf{Z}_t|\textbf{Z}_{t-1}, \textbf{h}_t))\\
+& \mathbb{E}_{\psi_t\sim q_{\phi}(\psi_t|\textbf{Z}_{t-1}, \textbf{a}_t)} \mathbb{E}_{\textbf{Z}_t\sim q(\textbf{Z}_t|\psi_t)} \log p(\textbf{A}_t|\textbf{Z}_t)\\
\end{aligned}
\end{equation}

\section{Experiments}
\subsection{Datasets}
SGRNN and SI-SGRNN are evaluated on six real-world dynamic graphs that are described in Table \ref{table_1}. And more detailed description can be found in supplements.

\begin{table}[t]
	\centering
	\caption{Datasets information}
	\scalebox{0.65}{
		\begin{tabular}{c|cccccc}
			\hline
			\textbf{Metrics}&Enron&Colab&Facebook&HEP-TH&Cora&Social Evolution\\
			\hline\hline
			Snapshots&11&10&9&40&11&27\\
			Nodes&184&315&663&1199-7623&708-2708&84\\
			Edges&115-266&165-308&844-1068&769-34941&406-5278&303-1172\\
			Average Density&0.01284&0.00514&0.00591&0.00117&0.00154&0.21740\\
			Node Attributes&-&-&-&-&1433&168\\
			\hline
	\end{tabular}}
	\label{table_1}	
\end{table}
\subsection{Baseline methods} The proposed methods are compared with six dynamic graph node embedding methods. \textbf{DynRNN} and \textbf{DynAERNN}\cite{goyal2020dyngraph2vec} are based on RNN models, which are used to prove the effectiveness of modeling the uncertainty in latent representations. Besides, \textbf{DynAE}\cite{goyal2020dyngraph2vec} is a deep autoencoder model with fully connected layers, which is used to compare the difference between MLP and RNN based sequential learning methods. \textbf{VGAE} and \textbf{GRNN}, as two basic components of the model, are also used for comparison. Note that VGAE is used for static graph so here it's implemented to analyze each snapshot separately. What's more, our methods are also compared with \textbf{VGRNN} and \textbf{SI-VGRNN} that model the uncertainty of latent representations. It's used to prove the advantage of structure of SGRNN that separates deterministic states from stochastic states. 
\subsection{Evaluation task} The definitions of \textit{link prediction} in dynamic graph are different in different papers, which is also stated in VGRNN. We evaluate SGRNN and its variants on three different \textit{link prediction} tasks that have been widely used in dynamic graph representation learning. Specifically, given a dynamic graph $\mathcal{G} = \{G^{(1)}, G^{(2)}, \cdots, G^{(T)}\}$ and its node attributes $\mathcal{X} = \{\textbf{X}_{1}, \textbf{X}_{2}, \cdots, \textbf{X}_{T}\}$, dynamic link prediction are defined as follows: (1) \textbf{Dynamic link detection} Detect unobserved edges in $G^{(t)}$. (2) \textbf{Dynamic link prediction} Predict edges in $G^{(t+1)}$. (3) \textbf{Dynamic new link prediction} Predict edges in $G^{(t+1)}$ that are not in $G^{(t)}$.
\begin{table*}[t]
	\centering
	\renewcommand\arraystretch{0.8}
	\small 
	\caption{Dynamic link detection}
	\begin{tabular}{c|c|cccccc}
		\hline
		\textbf{Metrics}&Methods&Enron&Colab&Facebook&HEP-TH&Cora&Social Evo.\\
		\hline\hline
		\multirow{9}*{AUC}&VGAE&88.26$\pm$1.33&70.49$\pm$6.46&88.37$\pm$0.12&79.31$\pm$1.97&87.60$\pm$0.54&79.85$\pm$0.85\\
		& DynAE&84.06$\pm$3.30&66.83$\pm$2.62&60.71$\pm$1.05&63.94$\pm$0.18&53.71$\pm$0.48&71.41$\pm$0.6\\
		& DynRNN&77.74$\pm$5.31&68.01$\pm$5.05&69.77$\pm$2.01&72.39$\pm$0.63&76.09$\pm$0.97&74.13$\pm$1.74\\
		& DynAERNN&91.71$\pm$0.94&77.38$\pm$3.84&81.71$\pm$1.51&82.01$\pm$0.49&74.35$\pm$0.85&78.67$\pm$1.07\\
		& GRNN&91.09$\pm$0.67&86.40$\pm$1.48&86.60$\pm$0.59&89.00$\pm$0.46&91.35$\pm$0.21&78.27$\pm$0.47\\
		& VGRNN&94.41$\pm$0.74&88.67$\pm$1.57&88.00$\pm$0.57&91.12$\pm$0.71&92.08$\pm$0.35&82.69$\pm$0.55\\
		&\textbf{SGRNN}&\textbf{96.81$\pm$0.43}&\textbf{89.66$\pm$0.48}&\textbf{89.34$\pm$0.23}&\textbf{94.25$\pm$0.83}&\textbf{93.93$\pm$0.28}&\textbf{83.61$\pm$0.47}\\
		&SI-VGRNN&95.03$\pm$1.07&89.15$\pm$1.31&88.12$\pm$0.83&91.05$\pm$0.92&94.07$\pm$0.44&83.36$\pm$0.53\\
		&\textbf{SI-SGRNN}&\textbf{97.63$\pm$0.98}&\textbf{89.96$\pm$1.27}&\textbf{89.31$\pm$0.71}&\textbf{94.31$\pm$1.02}&\textbf{95.24$\pm$0.39}&\textbf{84.32$\pm$0.51}\\
		\hline
		
		\multirow{9}*{AP}&VGAE&89.95$\pm$1.45&73.08$\pm$5.70&79.80$\pm$0.22&81.05$\pm$1.53&89.61$\pm$0.87&79.41$\pm$1.12\\
		& DynAE&86.30$\pm$2.43&67.92$\pm$2.43&60.83$\pm$0.94&63.87$\pm$0.21&53.84$\pm$0.51&70.18$\pm$1.98\\
		& DynRNN&81.85$\pm$4.44&73.12$\pm$3.15&70.63$\pm$1.75&74.12$\pm$0.76&76.54$\pm$0.66&72.15$\pm$2.30\\
		& DynAERNN&93.16$\pm$0.88&83.02$\pm$2.59&83.36$\pm$1.83&85.57$\pm$0.93&79.34$\pm$0.77&77.41$\pm$1.47\\
		& GRNN&93.47$\pm$0.35&88.21$\pm$1.35&84.77$\pm$0.62&89.50$\pm$0.42&91.37$\pm$0.27&76.93$\pm$0.35\\
		& VGRNN&95.17$\pm$0.41&89.74$\pm$1.31&87.32$\pm$0.60&91.35$\pm$0.77&92.92$\pm$0.28&81.41$\pm$0.53\\
		&\textbf{SGRNN}&\textbf{97.15$\pm$0.24}&\textbf{91.68$\pm$0.41}&\textbf{89.98$\pm$0.21}&\textbf{93.75$\pm$1.03}&\textbf{94.72$\pm$0.19}&\textbf{84.56$\pm$0.49}\\
		&SI-VGRNN&96.31$\pm$0.72&89.90$\pm$1.06&87.69$\pm$0.92&91.42$\pm$0.86&94.44$\pm$0.52&83.20$\pm$0.57\\
		&\textbf{SI-SGRNN}&\textbf{97.40$\pm$0.91}&\textbf{92.03$\pm$0.98}&\textbf{90.03$\pm$0.42}&\textbf{93.90$\pm$1.15}&\textbf{94.98$\pm$0.71}&\textbf{85.58$\pm$0.51}\\
		\hline
	\end{tabular}
	\label{table_2}
\end{table*}
\subsection{Experimental setups} 
Equal number of true and false edges are selected to evaluate the ability to distinguish them for performance comparison. Specifically, 5\% and 10\% of true edges at each snapshot and equal number of non-link node pairs are randomly selected for validation and test in dynamic link detection task. And for dynamic (new) link prediction, the (new) edges of next snapshot and equal number of non-link node pairs are randomly selected for evaluation. The last three snapshots of dynamic graphs are selected for test while the rest snapshots are used for parameter learning except for HEP-TH. We choose the last 10 snapshots for test for HEP-TH. Besides, average precision (AP) and area under the ROC curve (AUC) scores of link prediction are selected to evaluate performance of the methods. What's more, some datasets have no node attributes, so an $N_t$-dimensional identity matrix is selected as node attribute at time t. The mean and standard deviation for 10 runs are given in the corresponding tables.

The hidden states $h_t$ are computed with a single recurrent layer with 32 hidden dimensions. As for $a_t$, it's implemented with a GCN layer when the task is link detection while a fully-connected layer for the other two tasks. Two fully-connected layers are implemented for $\boldsymbol{\mu}_t^{(p)}$ and $\boldsymbol{\sigma}_t^{(p)}$ with sizes [32, 20]. And a 2-layer GCN is implemented to model $\boldsymbol{\mu}_t^{(q)}$ and $\boldsymbol{\sigma}_t^{(q)}$. For SI-SGRNN, a stochastic GCN layer with size 32 and an additional GCN layer with size 20 are implemented to model inference network. $\boldsymbol{\epsilon}_t$ is implemented with 20-dimensional Gaussian noise. And we choose $\gamma=0.8$ for all datasets. All the methods are trained for 1500 epochs with learning rate 0.01. Early stopping strategy is utilized based on validation performance. Our model is implemented with Pytorch\cite{paszke2019pytorch} and detail hyperparameter settings and implementation can be found at \url{https://github.com/StochasticGRNN/SGRNN}.

\subsection{Results and Discussion}

\subsubsection{Dynamic link detection.} 
$\textbf{u}_t$ is set as $\{\textbf{A}_{t}, \textbf{X}_{t}\}$ for unobserved link detection. Table \ref{table_2} illustrates the results for dynamic link detection and it's obvious that our methods outperforms the others on all six datasets. When compared to VGAE, it proves the importance of modeling temporal dependencies. Compared to GRNN and DynAERNN, it proves that stochastic latent variables contains more information and more robust than deterministic latent states. Besides, it also demonstrates that separate deterministic states from stochastic states in the iterative process can help better model the evolution of dynamic graph compared to VGRNN and SI-VGRNN. In addition, the results on Cora and Social-evolution prove the importance of taking node attributes into account in the model.

The results of SI-SGRNN prove that flexible variational distribution can improve model's performance compared to standard Gaussian distribution. Compared to SI-VGRNN, SI-SGRNN also performs better. Besides, compared to SGRNN, SI-SGRNN significantly improves the results on Cora and Social-evolution datasets which contain node attributes. However, in some datasets, the performance of SI-SGRNN is basically the same with SGRNN. The reason may be that Gaussian distribution can already represent the latent variables well so that the performance is not improved by changing the latent distributions.

\subsubsection{Dynamic link prediction}
\begin{table}[t]
	\centering
	\caption{Dynamic link prediction}
	\scalebox{0.8}{
		\begin{tabular}{c|c|cccc}
			\hline
			\textbf{Metrics}&Methods&Enron&Colab&Facebook&Social Evo.\\
			\hline\hline
			\multirow{7}*{AUC}
			& DynAE&74.22$\pm$0.74&63.14$\pm$1.30&56.06$\pm$0.29&65.50$\pm$1.66\\
			& DynRNN&86.41$\pm$1.36&75.70$\pm$1.09&73.18$\pm$0.60&71.37$\pm$0.72\\
			& DynAERNN&87.43$\pm$1.19&76.06$\pm$1.08&76.02$\pm$0.88&73.47$\pm$0.49\\
			& VGRNN&93.10$\pm$0.57&85.95$\pm$0.49&89.47$\pm$0.37&77.54$\pm$1.06\\
			&\textbf{SGRNN}&\textbf{94.35$\pm$0.37}&\textbf{88.67$\pm$0.43}&\textbf{90.12$\pm$0.23}&\textbf{78.12$\pm$0.83}\\
			&SI-VGRNN&93.93$\pm$1.03&85.45$\pm$0.91&90.94$\pm$0.37&77.84$\pm$0.79\\
			&\textbf{SI-SGRNN}&\textbf{94.50$\pm$0.98}&\textbf{89.02$\pm$1.03}&\textbf{90.98$\pm$0.61}&\textbf{78.41$\pm$0.97}\\
			\hline
			
			\multirow{7}*{AP}
			& DynAE&76.00$\pm$0.77&64.02$\pm$1.08&56.04$\pm$1.37&63.66$\pm$2.27\\
			& DynRNN&85.61$\pm$1.46&78.95$\pm$1.55&75.88$\pm$0.42&69.02$\pm$1.71\\
			& DynAERNN&89.37$\pm$1.17&81.84$\pm$0.89&78.55$\pm$0.73&71.19$\pm$0.81\\
			& VGRNN&93.29$\pm$0.69&87.74$\pm$0.79&89.04$\pm$0.33&77.03$\pm$0.83\\
			&\textbf{SGRNN}&\textbf{94.58$\pm$0.24}&\textbf{89.92$\pm$0.57}&\textbf{89.37$\pm$0.36}&\textbf{78.01$\pm$0.69}\\
			&SI-VGRNN&94.44$\pm$0.85&88.36$\pm$0.73&90.19$\pm$0.27&77.40$\pm$0.43\\
			&\textbf{SI-SGRNN}&\textbf{94.96$\pm$0.91}&\textbf{90.87$\pm$0.80}&\textbf{90.03$\pm$0.42}&\textbf{78.19$\pm$0.81}\\
			\hline
	\end{tabular}}
	\label{table_3}
\end{table}
\begin{table}[t]
	\centering
	\caption{Dynamic new link prediction}
	\scalebox{0.8}{
		\begin{tabular}{c|c|cccc}
			\hline
			\textbf{Metrics}&Methods&Enron&Colab&Facebook&Social Evo.\\
			\hline\hline
			\multirow{7}*{AUC}
			& DynAE&66.10$\pm$0.71&58.14$\pm$1.16&54.62$\pm$0.22&55.25$\pm$1.34\\
			& DynRNN&83.20$\pm$1.01&71.71$\pm$0.73&73.32$\pm$0.60&65.69$\pm$3.11\\
			& DynAERNN&83.77$\pm$1.65&71.99$\pm$1.04&76.35$\pm$0.50&66.61$\pm$2.18\\
			& VGRNN&88.43$\pm$0.75&77.09$\pm$0.23&87.20$\pm$0.43&75.00$\pm$0.97\\
			&\textbf{SGRNN}&\textbf{90.38$\pm$0.59}&\textbf{81.49$\pm$0.32}&\textbf{87.71$\pm$0.49}&\textbf{77.12$\pm$0.83}\\
			&SI\_VGRNN&88.60$\pm$0.95&77.95$\pm$0.41&87.74$\pm$0.53&76.45$\pm$1.19\\
			&\textbf{SI-SGRNN}&\textbf{90.46$\pm$0.98}&\textbf{82.04$\pm$0.53}&\textbf{88.02$\pm$0.57}&\textbf{77.89$\pm$1.04}\\
			\hline
			
			\multirow{7}*{AP}
			& DynAE&66.50$\pm$1.12&58.82$\pm$1.06&54.57$\pm$0.20&54.05$\pm$1.63\\
			& DynRNN&80.96$\pm$1.37&75.34$\pm$0.67&75.52$\pm$0.50&63.47$\pm$2.70\\
			& DynAERNN&85.16$\pm$1.04&77.68$\pm$0.66&78.70$\pm$0.44&65.03$\pm$1.74\\
			& VGRNN&87.57$\pm$0.57&79.63$\pm$0.94&86.30$\pm$0.29&73.48$\pm$1.11\\
			&\textbf{SGRNN}&\textbf{90.19$\pm$0.32}&\textbf{82.81$\pm$0.40}&\textbf{86.98$\pm$0.34}&\textbf{75.05$\pm$1.03}\\
			&SI-VGRNN&87.88$\pm$0.84&81.26$\pm$0.38&86.72$\pm$0.54&73.85$\pm$1.33\\
			&\textbf{SI-SGRNN}&\textbf{90.21$\pm$0.91}&\textbf{83.79$\pm$0.72}&\textbf{87.07$\pm$0.58}&\textbf{75.62$\pm$1.22}\\
			\hline
	\end{tabular}}
	\label{table_4}
\end{table}
$u\textbf{}_t$ is set as $\{\textbf{A}_{t-1}, \textbf{X}_{t-1}\}$ for dynamic (new) link prediction in the next snapshot. And it should be noted that all the methods can not predict new nodes. Therefore, Cora and Hep-TH is not considered in this part. The results for dynamic (new) link prediction are shown in Table \ref{table_3} and \ref{table_4}. It's obvious that SGRNN and SI-SGRNN outperform other competing methods, especially on Colab and Enron. It indicates that our methods have better generalization because it introduces stochastic latent representations which provides more information and improves robustness. Besides, it separates deterministic states from stochastic states in the recurrent process and suppresses the interference between them. However, the performance on Facebook dataset is basically equivalent to that of VGRNN and SI-VGRNN. The reason could be that the stochastic states have little influence on deterministic states for representation in the recurrent process in (SI-)VGRNN on this dataset. 

In order to further analyze the capabilities of the model, we review the datasets carefully and find that Enron shows highest density and clustering coefficients. Therefore, many competing methods could get good results on this datasets. However, for the remaining three datasets, they have low density and clustering coefficients which make link prediction more difficult. And the results also demonstrate this point that many methods have bad performance. However, our methods can well deal with this problem and get best results around all competing methods, especially for dynamic new link prediction problem. 

Besides, the results of SI-SGRNN are basically the same with those of SGRNN in these two tasks. The reason may be that although the posterior distribution is more flexible, the prior is simply designed as Gaussian distribution. Therefore, there are several possible methods to improve the performance of the model. The first is to design a more flexible prior distribution. In addition, generative adversarial networks (GAN) could be introduced where our methods could be considered as a generator. However, GAN is difficult to converge and the training process needs to be carefully designed, which we leave for future studies.

\subsubsection{KL-vanishing}\label{sec_kl}
\begin{figure}[t]
	\centering
	\includegraphics[width=0.47\textwidth,height=0.22\textheight]{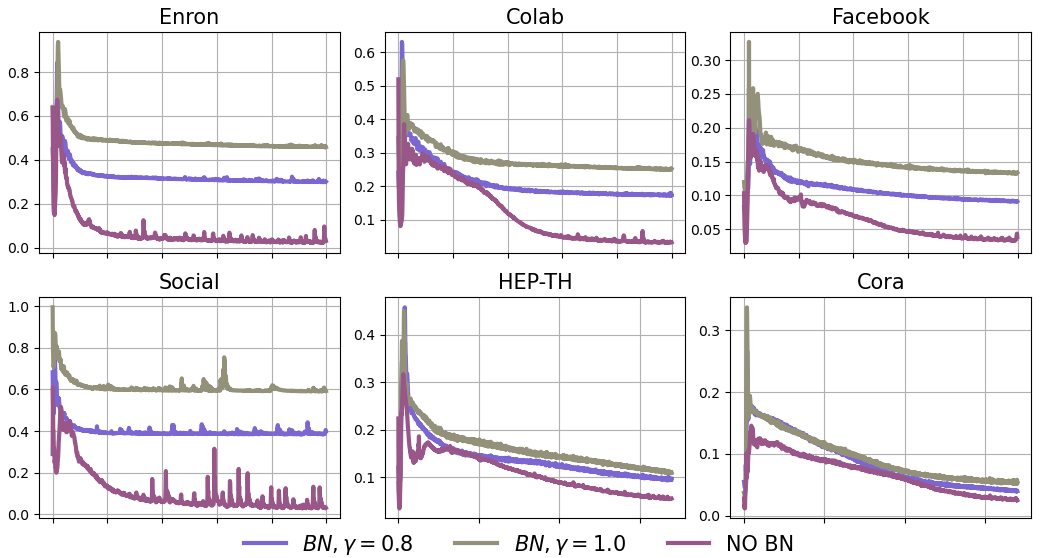} 
	\caption{Comparison of KL-divergence($\gamma=0.8$)}
	\label{kl-div}
\end{figure}

In our experiments, we found that SGRNN encountered KL-vanishing problem, which severely affected the expressive ability of the model. 
We compare the KL divergence of SGRNN with or without batch normalization on the six datasets and the results are shown in Fig \ref{kl-div}. The blue and red curves correspond to the change of KL divergence with the number of iterations when gamma is 1 and 0.8, respectively. And the green curve corresponds to the original SGRNN without batch normalization. The results demonstrate that the strategy is simple and efficient. What's more, it also demonstrates that the lower bound of KL-divergence can be controlled by adjusting $\gamma$. When $\gamma$ becomes larger, the lower bound also becomes larger, which is consistent with our derivation in section \ref{sec_KL_a}. 

\begin{figure}[t]
	\centering
	\includegraphics[width=0.44	\textwidth, height=0.22\textheight]{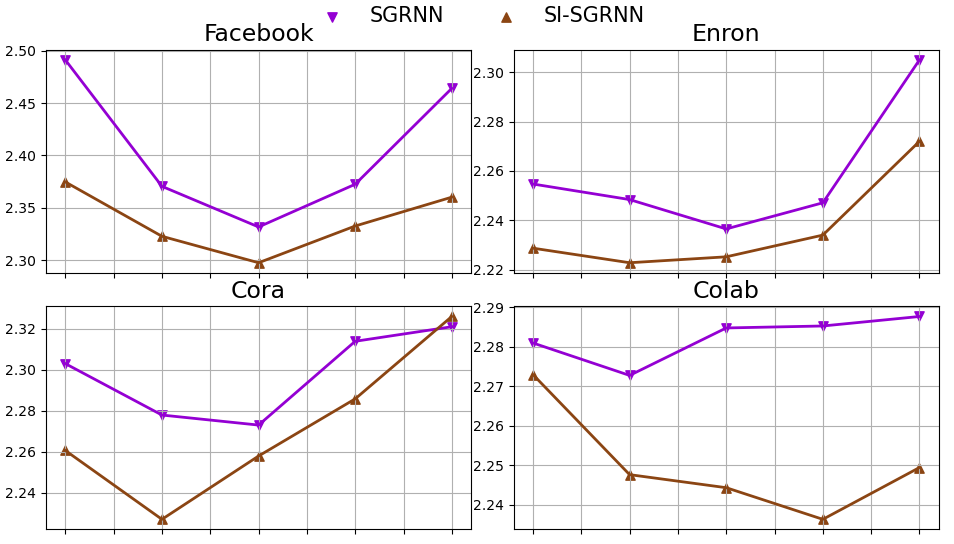}
	\caption{Negative log-likelihood for different $\gamma$.}
	\label{nll}
\end{figure}
In addition, four datasets are selected to calculate the sum of mean value of negative log likelihood at each snapshot on the test datasets for different $\gamma$ and the results are illustrated in Fig \ref{nll}. The range of $\gamma$ is from 0.6 to 1.0. It can be known from Fig \ref{nll} that the lower bound of KL-divergence can affect the negative log-likelihood. The negative log-likelihood first becomes smaller and then increases as $\gamma$ becomes larger. Therefore, the values of $\gamma$ corresponding to the area with smaller negative log likelihood can be used as candidates for training. Besides, SI-SGRNN can achieve lower negative log-likelihood compared with SGRNN in most cases, which prove that flexible variational distribution can enhance the performance of the model once again.

\begin{figure}[t]
	\centering
	\includegraphics[width=0.48\textwidth, height=0.18\textheight]{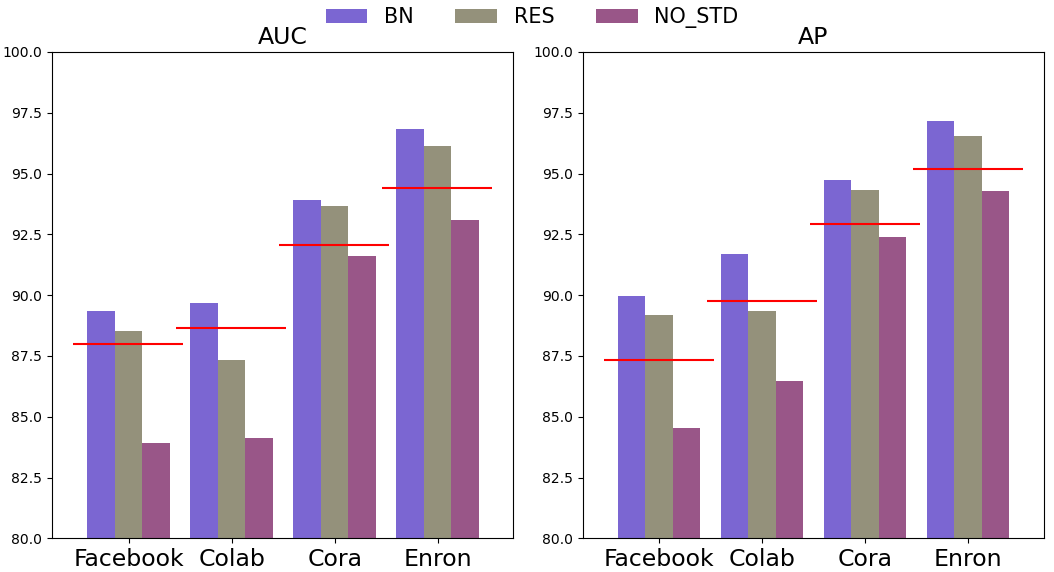} 
	\caption{Comparison of fixed BN with other two variants}
	\label{vakl}
\end{figure}

Two variants of Eq. \ref{stru} are integrated into SGRNN for comparison and the results are shown in Fig \ref{vakl}. The red horizontal lines represent the performance of VGRNN. RES represents $\boldsymbol{\mu}_q = \boldsymbol{\mu}_p + \text{NN}_1^{(q)}(\textbf{Z}_{t-1}, \textbf{a}_t)$. It shows that direct optimization of the difference between the prior mean and posterior mean can also achieve good results and performs better than VGRNN on three datasets. NO\_STD represents $\boldsymbol{\mu}_q = \boldsymbol{\mu}_p + \text{FixedBN}(\text{NN}_1^{(q)}(\textbf{Z}_{t-1}, \textbf{a}_t))$. It's used to demonstrate that batch normalization should be used for $(\boldsymbol{\mu}_q -\boldsymbol{\mu}_p)/\boldsymbol{\sigma_p}$ instead of  $\boldsymbol{\mu}_q -\boldsymbol{\mu}_p$.

\begin{figure}[t]
	\centering
	\includegraphics[width=0.48\textwidth,  height=0.18\textheight]{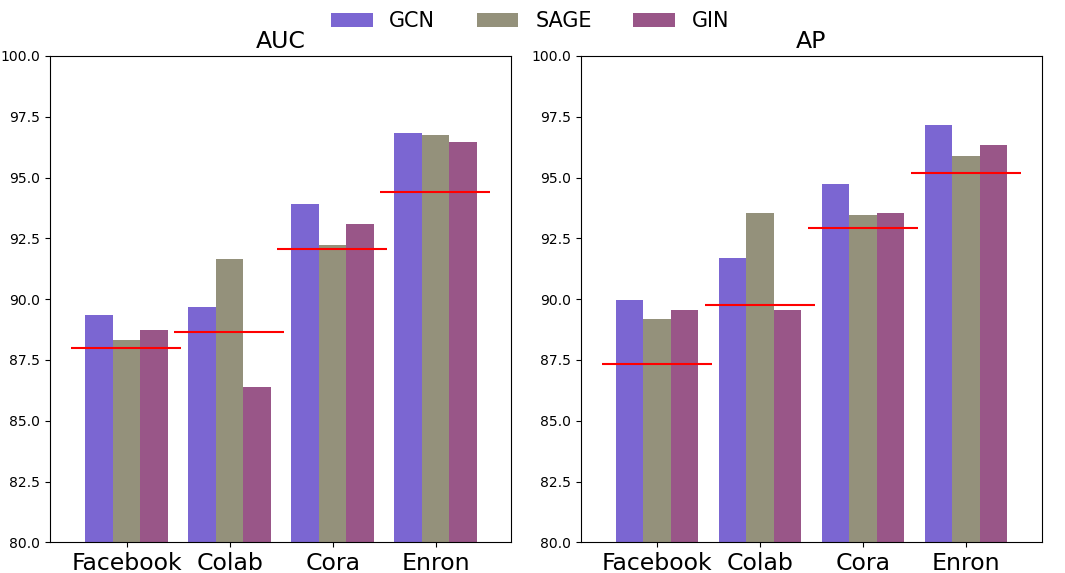} 
	\caption{The effect of GNN type on model performance}
	\label{gnn}
\end{figure}

The influence of different types of graph neural networks on the performance of SGRNN is also explored . GCN, GraphSAGE and GIN\cite{xu2018powerful} are selected and the results are shown in Fig \ref{gnn}. The red horizontal lines represent the performance of VGRNN that implemented with GCN. It shows that SGRNN performs better than VGRNN in most cases regardless of GNN types, which further prove the efficiency of SGRNN. Besides, SGRNN implemented with GCN performs better compared to GraphSAGE and GIN in most cases. However, SGRNN implemented with GraphSAGE performs best on Colab. In conclusion, the type of GNN has a influence on model performance in some way, especially on sparse graph.

\section{Conclusion}

In this paper, we propose SGRNN and SI-SGRNN for the representation learning over dynamic graphs, which model the uncertainty of latent representations and capture the changes in node attributes and graph topology simultaneously. Besides, the deterministic states and stochastic states are separated in the recurrent process for mutual interference suppression. To alleviate KL-vanishing problem, a novel structure for posterior approximation is proposed based on the lower bound of KL divergence, which can be simply implemented with one line code modification. And experiments on six real word datasets demonstrates the efficiency of the methods. Besides, the influence of $\gamma$ in batch normalization and different types of GNN on model expressive ability is explored. In the future, more flexible prior distribution and GAN can be introduced to possibly further improve model expressive ability.

\bibliography{ref}
\bibliographystyle{IEEEtran}
\end{document}